\documentclass[12pt]{elsarticle}
\usepackage{multirow}

\usepackage{amssymb}
\usepackage{amsthm}

\usepackage[left=2.5cm, right=2.5cm, top=4cm, bottom=3cm, footskip=0.5cm]{geometry}

\journal{ }

\begin{document}

\begin{frontmatter}

\title{Exploring the Sensory Spaces of English Perceptual Verbs in Natural Language Data}
\date{ }

\author[1,2]{Roxana Girju}
\author[2,*]{David Peng}

\address[1]{Department of Linguistics and Beckman Institute, University of Illinois at Urbana-Champaign, Urbana IL 61802}
\address[2]{Department of Computer Science, Department of Economics, University of Illinois at Urbana-Champaign, Urbana IL 61802}
\address[*]{Corresponding author: girju@illinois.edu}

\begin{abstract}
In this study, we explore how language captures the meaning of words, in particular meaning related to sensory experiences learned from statistical distributions across texts. We focus on the most frequent perception verbs of English analyzed from an an Agentive vs. Experiential distinction across the five basic
sensory modalities: Visual ({\tt to look} vs. {\tt to see}), Auditory ({\tt to listen} vs. {\tt to hear}), Tactile ({\tt to touch} vs. {\tt to feel}), Olfactory ({\tt to smell}), and Gustatory ({\tt to taste}). 
In this study we report on a data-driven approach based on distributional-semantic word embeddings and clustering models to identify and uncover the descriptors’ sensory spaces of the perception verbs. In the analysis, we identified differences and similarities of the generated descriptors based on qualitative and quantitative differences of the perceptual experience they denote. For instance, our results show that while the perceptual spaces of the experiential verbs like {\tt to see, hear} show a more detached, logical way of knowing and learning, their agentive counterparts ({\tt to look, listen}) provide a more intentional as well as more intimate and intuitive way of discovering and interacting with the world around us.
We believe that such an approach has a high potential to expand our understanding and the applicability of such sensory spaces to different fields of social and cultural analysis. Research on the semantic organization of sensory spaces for various applications might benefit from an the Agentive/Experiential account to address the complexity of multiple senses wired with each other in still unexplored ways.
\end{abstract}


\end{frontmatter}


\section{Introduction}
Embodied cognition is a growing paradigm within cognitive science that stipulates that cognitive processes are deeply rooted in the body’s interactions with the world \cite{Farina2020}. Here, concepts are viewed as mental structures that generalize over experiences, being grounded in the sensory and motor contexts of their occurrence. Thus, concept acquisition can happen through direct (experience of self) or indirect (observing the experience of others) sensory-motor experience, as well as through language. 

It is agreed that language can encode, to some extent, the perception system of sensory and bodily experience \cite{Louwerse2010}. However, it is still unclear how are these sensory perceptions captured in language and which perceptual qualities are easier to express verbally than others \cite{Winter2019}.
Researchers have analyzed word distributions in various sensory modalities \cite{SanRoque2015} in literal and figurative contexts and across languages \cite{Sanroque2018}. Yet, to the best of our knowledge, no research has taken steps toward an integrated look along various  distinctions across the five basic perception modalities, and the semantic organization of these sensory spaces with a large-scale text data-driven model.

In this study, we investigate how the English language captures the meaning of words and the concepts they encode, and how meaning related to sensory experiences is learned from statistical distributions across large English texts. In doing so, we focus on the most frequent English perception verbs analyzed along two main dimensions: (1) from an Agentive/Experiential distinction, and (2) across the five basic perception modalities: Visual ({\tt to look} vs. {\tt to see}), Auditory ({\tt to listen} vs. {\tt to hear}), Tactile ({\tt to touch} vs. {\tt to feel}), Olfactory ({\tt to smell}), and Gustatory ({\tt to taste}). While it is well accepted that the verb {\tt to feel}, for instance, captures a much broader and nuanced experiential domain outside its primary area of tactile perception, it is still unclear which experiences mix in and how much of the sensory-motor experience is captured in the mental representation \cite{Divjak2015,PengGirju2021}. 

It is important to note that the Agentive/Experiential distinction does not refer to syntactic properties of the perception verbs considered. An agent is an entity that intentionally carries out the action of the verb, while an experiencer undergoes an emotion, a state of being, or a perception expressed by the verb \cite{Levin1993}. When  it comes to the Agentive/Experiential distinction, in this paper, we slightly depart from the precedent in the field  which is based on Viberg's classification \cite{Viberg1984,SanRoque2015,Sanroque2018}. In his seminal paper on the verbs of perception, Viberg distinguishes between subject-oriented verbs where the grammatical subject is the perceiver, and object-oriented verbs where the grammatical subject is the stimulus of perception. The first category includes agentive (i.e., "Anna looked at the sculpture") and experiencer (i.e., "Anna saw the sculpture") acts of perception. In the second class, Viberg includes intransitive verbs where the perceiver does not need to be expressed linguistically at all. One such example is "It looks funny", where instead of focusing on the act of perception itself, the verb tends to indicate an assessment or value judgment toward the object of perception. In this research we are particularly interested in the Agentive/Experiential divide, and thus, do not explicitly encode object-oriented verbs. The reasons are two fold: (1) since we deal with a large dataset, it is impractical to grammatically process the text to identify different types of grammatical subjects; and (2) our data-driven word embedding model is able, to some extent, to capture such class of verbs automatically based on the context that defines it. For instance, our model could automatically capture differences between forms like {\tt to sniff, to smell}, as in "I sniffed the cheese" vs. "The cheese smells like dirty socks".\footnote{The authors like to thank the anonymous reviewer who pointed this aspect out and who suggested these examples.}

Although there have been attempts at exploring the grammar of such verbs of perception \cite{Divjak2015}, we are not aware of any large-scale descriptive system that has tried to organize the wide and complex diversity of their sensory spaces. 
We employ here a large data-driven approach to identifying and organizing the descriptor space of the main eight perceptual verbs based on qualitative and quantitative differences of the sensory experience they denote. Specifically, our aims are: 
\begin{enumerate}
    \item to explore to which extent and along what detailed aspects our distributional approach helps in acquiring the vast and diverse set of contextual descriptors related to the main English perception verbs; and
    \item to analyze their resulting conceptual and sensory spaces.
\end{enumerate}   

In this study, we explore the power of language to analyze nuances of sensory descriptors. In addressing these issues, we consider a large-scale text data-driven framework based on a distributional-semantic word embedding model and clustering algorithm to identify and uncover the descriptors’ sensory spaces. We analyze the generated sensory descriptors along the two dimensions of interest in our dataset and compare the results with the Lancaster Sensorimotor Norms project \cite{Lynott2020}, to date the largest set of English semantic sensory norms. 

We believe that such an approach has a high potential to expand our understanding and the applicability of such sensory spaces to different fields of social and cultural analysis. Research on the semantic organization of sensory spaces for various applications might benefit from an the Agentive/Experiential account to address the complexity of multiple senses wired with each other in still unexplored ways, and to understand their prototypical and periferic placement across material and immaterial worlds. 
Moreover, the findings are relevant for research on concept acquisition and representation, as well as for any application that can benefit from a better understanding of perceptual spaces of sensory experiences.

\section{Methodological Approach}

In this paper, we focus on the three of the five basic sense modalities as expressed by the most frequent perception verbs of English in Agentive/Experiential forms -- i.e., Visual ({\tt to look} vs. {\tt to see}), Tactile ({\tt to touch} vs. {\tt to feel}), and Auditory ({\tt to listen} vs. {\tt to hear}) - as they categorize the sensory events at the basic level of categorization. Separately, we considered the verbs {\tt  smell} and {\tt taste} to represent the Olfactory and Gustatory modalities, respectively, even though these do not seem to have an Agentive/Experiential split.

We explored and report here our findings on the sensory spaces automatically extracted for these perception verbs along two dimensions: (1) the five perception modalities (Vision, Auditory, Tactile, Olfactory, and Gustatory); and (2) the Agentive/Experienetial modes.\footnote{Our code and collocational data are made available on GitHub (https://github.com/davidtpeng/sensory-spaces-perceptual-verbs).}
Below we explain the data collection and processing, the computational modeling of the sensory spaces across the dimensions of interest, and evaluate and interpret the results.

\subsection{Creating the Corpus}

Sentences containing the inflections of all the perception verbs considered were extracted from the University of Maryland Baltimore County (UMBC) web-based collection \cite{UMBC}, a dataset of English paragraphs with over three billion words processed from the February 2007 crawl from the Stanford WebBase project (http://bit.ly/WebBase). 
The part-of-speech (POS) tagged version of the corpus (following the Penn Treebank tokenization style \cite{marcus-etal-1994-penn})
allowed us to distinguish among different homonyms of various lexical categories.
We pre-processed the corpus by converting all words to lower case, lemmatizing content words, and returning one sentence per line. From this pre-processed corpus, we extracted a sub-corpus (henceforth the perception corpus) corresponding to roughly 1/8th of the original corpus\footnote{This subcorpus corresponds to the first 50 json files of the UMBC corpus.} sentences which contained the corresponding verbs and their inflections.
Table~\ref{corpus-stats} shows the distribution of each perception verb across the two Agentive/Experiential split.

\begin{table}
\centering
\begin{tabular}{lcr}
\hline \textbf{Sensory} & \textbf{Agentive/Experiential} & \textbf{Total} \\
\textbf{modality} & \textbf{Distinction} &   \\ \hline
Visual &  & 642,256 \\ 
      & \multicolumn{1}{l}{to look} & 244,815   \\
      & \multicolumn{1}{l}{to see}  & 397,441 \\ \cline{2-3}
Auditory & & 166,177 \\ 
        & \multicolumn{1}{l}{to listen} & 37,321  \\
        & \multicolumn{1}{l}{to hear} & 128,856  \\ \cline{2-3}
Tactile & & 206,283 \\ 
      & \multicolumn{1}{l}{to touch} & 27,648  \\
      & \multicolumn{1}{l}{to feel}  & 178,635 \\ \cline{2-3}
Gustatory &  &  17,676 \\ \cline{2-3}
Olfactory &  &  7,897 \\
\hline
\end{tabular}
\caption{\label{corpus-stats} Number of sentences per perceptual verb, sensory modality, and agentive/experiential type in the perception corpus.}
\end{table}

\subsection{Identifying Perceptual Descriptors}

In this subsection we show how we identified and extracted sensory spaces - i.e., the descriptor words associated with our eight perception verbs of interest. A quick literature review on sensory descriptors shows that there have been very few recent studies that have investigated the semantic and perceptual space on the basis of natural language data --  mostly on taste and odors \cite{Iatropoulos-etal2018,McGregorM16,Horberg-etal2020}.
Other research which did or did not take advantage of the computational capabilities offered by the field of natural language processing, has generated rather limited, out of context sets of pre-selected descriptors, classified into categories, or ranked based on various descriptor rating scores (see \cite{Lynott2020} for details).
In a study of the semantic content of olfactory words in a large English text corpus \cite{Iatropoulos-etal2018}, for example, Iatropoulos and colleagues introduced the Olfactory Association Index (OAI), a metric that captures how strongly a word is associated with the concept of smell. The OAI metric was validated on psychophysical datasets showing that concepts with high OAI have high ratings of perceived olfactory association and thus can be used to describe highly familiar odors. The OAI is the $log_{2}$ probability that a descriptor $d$ occurs in olfactory-related as opposed to non-olfactory contexts: 
\begin{equation}
OAI_{d} = log_{2} {of}_{d} / {tf}_{d}, 
\end{equation}
where ${tf}_d$ is the total frequency of $d$ and ${of}_{d}$ is the frequency of $d$ in olfactory-related contexts.
These researchers, however, focused on identifying only descriptors of taste and smell and did not show how those descriptors are organized semantically across multiple sensory modalities. 

In our present work, we go beyond such studies along the dimensions identified above. Specifically, 
from the perception corpus, we extracted all the descriptors -- i.e., lemma words with parts of speech identifying content words: nouns, verbs, adjectives, and adverbs -- that occurred at least 10 times within a +/-4 word window of each perception verb. While the frequency of occurrence of 10 and word context window of +/-4 are common parameter values used to extract sensory descriptor lists \cite{Horberg-etal2020}, they should be definitely tested and validated for various data sets and the situational context of the application domain. However, since such an analysis requires considerable compute power as well as time, we will empirically determine the best value of these parameters in future research extensions of our work.

From the extracted context windows, for each verb we selected the top 1,600 most frequently occurring descriptors and ranked them according to the OAI metric which was calculated on the basis of modality-centric frequency and total frequency, and extracted their corresponding lemmas, parts of speech, and word forms. The cutoff value of maximum 1,600 was determined by manual inspection -- meaning, when the descriptors began to become irrelevant or noisy. However, this cutoff value was not the same for all the modality descriptor lists. For instance, the cutoff value for the descriptors associated with the verb {\tt to see} was set to 1,400 given noisy words like "k.s.a", "oracle9i", "mpep" showing after that cutoff. As Table~\ref{corpus-stats} shows, some perception verbs like {\tt to smell} had fewer instances in the corpus, 
much less than 1,600 since they occurred less frequently. However, for consistency purposes, we decided to keep all descriptors that occur at least 8 times to be considered for the OAI score calculation. Most of the extracted descriptor lists have a little less than 1,600 descriptors.

Since, in our study we use this metric beyond the olfactory to reflect information about all five basic modalities, we refer to it as \emph{Perception Association Index}, henceforth PAI. 
Given that the descriptor list sizes vary per sensory verb, we then selected for clustering the top 300 descriptors that had the highest PAI scores. For the remainder of the paper, we refer to these as sensory descriptors or concepts.

\begin{figure}[h!b]
  \caption{\label{POS300} The distribution of the top 300 descriptors' parts of speech per each sensory modality and perception verb: NN 
  (nouns), JJ (adjectives), VB (verbs), RB (adverbs), and MD (modals).}
  \centering
    \includegraphics[width=0.7\textwidth]{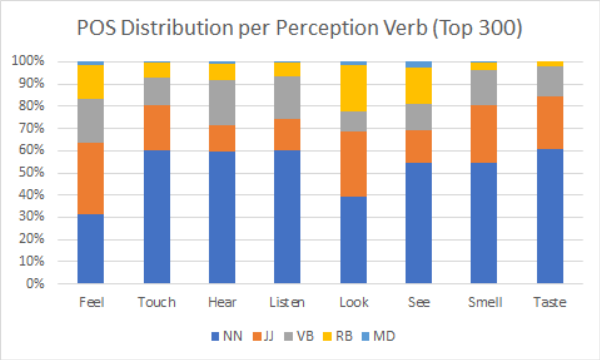}
\end{figure}

Figure \ref{POS300} shows the part of speech distribution of the top 300 sensory descriptors per each sensory modality and perception verb. We can see that while nouns predominate in most of the sensory contexts, adjectives are more prevalent in the contexts of {\tt to feel, look, smell}, and {\tt touch}, while adverbs occur more with verbs {\tt to hear, listen, feel} and not so much with the verbs {\tt to look} or {\tt see}. Modals are insignificant in number across all the perceptual verb contexts, and thus we did not consider them for further processing.

\section{Computational Model}
Taking advantage of the concept distributions in the specific sensory contexts around the perception verbs, we automatically identified sensory descriptors by determining the extent to which they express specific sensory-semantic content. We then derived their semantic organization using a distributional-semantic word embeddings model. In such a model, words are represented as vectors in a multi-dimensional space. Our model allows us to classify words according to the perception type, and then chart the similarities and differences in the verbs' preference for contextual words. 


We trained the distributional-semantic word embedding model (word2vec with CBOW) \cite{mikolov2013} on the entire UMBC billion corpus 
containing our perceptual verbs using a +/-4 word context-window, a hidden layer of 200 units, a minimum word frequency count of 10, and 30 training iterations. This resulted in a multi-dimensional space of word vectors where the semantic differences between words are represented as vector distances. We extracted concept vectors from our language model corresponding to the selected 300 top sensory descriptors per perception verb. Each concept vector was standardized and used to derive the descriptor distance matrices.
 The distance $D$ between descriptors $i$ and $j$ was calculated as 
 \begin{equation}
    {D}_{ij} = 0.5 * (1 - {p}_{ij}),
 \end{equation} 
 where $p$ is the Pearson correlation between concept vectors. As such, the distance $D$ between descriptors is converted to the 0--1 range, with 0 reflecting semantic identity and 1 indicating semantic opposition. 
 
The distance matrices were analyzed with Agglomerative Nesting (AGNES) hierarchical clustering \cite{AGNES}, a bottom-up clustering method that successively merges clusters based on their distance, quantified with Ward's minimum variance method \cite{Ward1963}.
The clustering model generated descriptor clusters (visualized as dendograms) per the primary dimensions of interest along which the descriptors are compared. 

It is important to note that, in clustering, the interpretation of such semantic partitioning is done manually by the human analyst.
There is no correct or incorrect number of clusters to define - it really depends on what the user expects a “natural” grouping to be. A clustering outcome can be measured along two dimensions: compactness (to what extent data points within each cluster are similar to each other), and closeness (to what extent data points grouped together are closer to each other than they are to observations in other groups).
In order to interpret and validate the consistency within clusters of data, we performed a silhouette analysis. Silhouette analysis \cite{Rousseaew1987} is the average of the ratio of each cluster’s compactness and closeness. For agglomerative hierarchical clustering, a silhouette coefficient can be computed for several cuts (k = 2 \dots N-1) and the user selects the k value with the maximum silhouette coefficient. 
This measure ranges from -1 to +1, where a high value indicates that the data point is well matched to its own cluster and poorly matched to neighboring clusters. If many points have a low or negative value, then the clustering configuration may have too many or too few clusters.

\begin{table} [htbp]
\centering
\small
\begin{tabular}{lrr} \hline 
\textbf{Descriptor} & \textbf{Max Silhouette} & \textbf{Optimal No.} \\ 
\textbf{Set} & \textbf{Score} & \textbf{of Clusters} \\ \hline
Feel (Purple Cluster) &	0.201 &	5 \\
Feel (Top 300 descriptors by PAI) &	0.1 & 105\\
Hear (Top 300 descriptors by PAI) &	0.407 &	2\\
Look (Top 300 descriptors by PAI) &	0.189 &	2\\
See (Top 300 descriptors by PAI) &	0.252 &	2\\
Smell (Top 300 descriptors by PAI) &	0.143 &	2\\
Taste (Top 300 descriptors by PAI) &	0.187 &	2\\
Touch (Top 300 descriptors by PAI) &	0.113 &	2\\
Listen (Top 300 descriptors by PAI)	& 0.102	& 74\\
\hline
\end{tabular}
\caption{\label{silhouette} Results of the silhouette analysis to evaluate the quality of the generated descriptor clusters. The Purple Cluster is the largest cluster for the verb {\tt to feel} (55\% of all {\tt feel} descriptors) on which we ran hierarchical clustering again (see Section 4.3 for details).
}
\end{table}

Table~\ref{silhouette} gives the silhouette scores for our perception verbs which averaged around 0.186, ranging from 0.1 for {\tt to feel} to a notable outlier of 0.407 for {\tt to hear}. These scores indicate an overall modest separation between perception verb clusters. The verb {\tt to hear} has a high silhouette score as its descriptors can be neatly separated into two clusters: descriptors related to sound (i.e., \emph{gunshot, listener, radio}) and descriptors related to trial proceedings (i.e., \emph{transcript, judge, testify}). Additionally, it is important to note that when we zoom into specific subclusters, the silhouette score increases. For instance, the descriptors in the large purple cluster focusing on emotion (170 descriptors) had a silhouette score of 0.201 vs. 0.1 for all {\tt feel} descriptors, and the optimal number of clusters drops from 105 to 5. This indicates that subclusters may be more coherent than all descriptors as a whole. 


In the future we will consider other methods as well, such as the Gap Statistic \cite{GapStatistic2001} and methods based on sum-of-square measurements.
In practice, however, the best way to evaluate your clustering results is to start by actually examining manually the clusters formed and making a determination based on an understanding of what the data represents, what a cluster represents, and what the clustering is intended to achieve. This is particularly important for our analysis here given the cross-sensory nature of the spaces generated by the perception verbs (see the dendogram generated for the descriptors of {\tt touch} in Figure \ref{touch-dandogram}).

\section{Results of the Clustering Model and their Analysis}
\label{sec:length}

\begin{figure}[htbp]
  \caption{\label{heatmap} The sensory heatmap for the number of descriptors in common for each pair of 300 concepts.}
  \centering
    \includegraphics[width=0.7\textwidth]{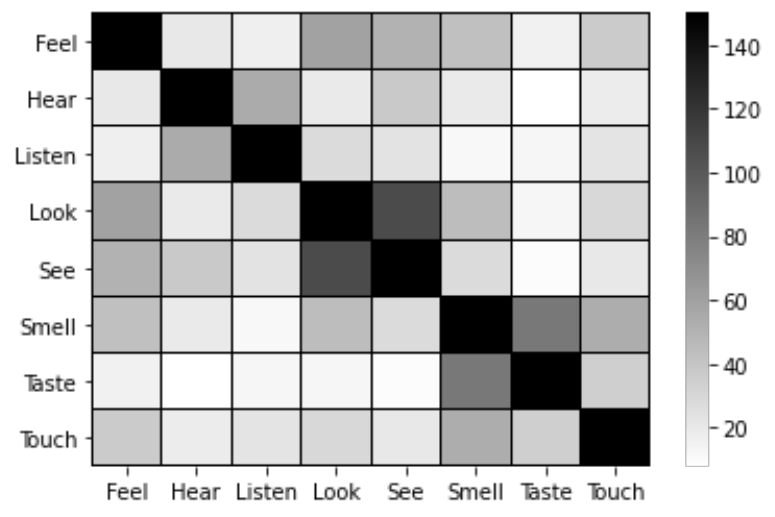}
\end{figure}

In this section we present and analyze in detail the results of the clustering model and compare the spaces of sensory descriptors along two points of interest: (1) the five basic sensory modalities and (2) the Agentive/Experiential perception verb split.
To help us in this process, we built a heatmap diagram based on the pairwise associations of the 300 descriptors of each sensory verb -- meaning, we calculated the number of descriptors in common for each pair of sets of 300 concepts. The heatmap diagram in shown in Figure \ref{heatmap} and is analyzed in the following subsections along the two dimensions of interest.

\subsection{The Visual Modality: \emph{to Look} vs. \emph{to See}}
As the heatmap diagram in Figure \ref{heatmap} shows, the two perception verbs with the highest overlap of their sensory spaces are the two verbs of Vision sharing one third of the descriptors. The intersection captures similarities of direction of sight (i.e., \emph{forward, around, ahead}), facial parts and features (\emph{face, eyes, smile}), objects of observation (\emph{mirror, window, glass, photo, painting, scene}) as well as a high number of generic concepts like \emph{people, folk, color}, and \emph{elsewhere, everywhere, somebody, anybody}.

The clusters captured by the generated dandograms for the sensory descriptors of {\tt to look} and {\tt to see} varied in five interesting aspect categories. The first category is seen across modifiers like adjectives and adverbs. While the sensory space of {\tt to see} captures words like \emph{clarity, clearly, obviously}, that of {\tt to look} is seen with many descriptors like \emph{closely, carefully, briefly, specifically, in depth}, indicating a deeper attention to detail and zoom in effect that shows not only a higher level of agentivity, but also a shorter proximity to the object in sight. Static/dynamic is another interesting category where we see clear differences. While the agentive verb of vision ({\tt to look}) identifies static objects of observation and detail like \emph{landscapes, telescopes, microscopes}, the experiential counterpart ({\tt to see}) is seen more often with dynamic objects like \emph{TV, screen, movies, film}, as well as descriptors like \emph{curious, surprised}, and \emph{exciting}. The two vision verbs also vary considerably across the descriptor space when it comes to who is doing the seeing/looking. {\tt To look} is more often associated with subjects like \emph{buyers, clients, investors} which take more of an agentive role than \emph{observers} and \emph{viewers} which are seen more with the verb {\tt to see}. The two sight verbs also seem to differ in their selection of color words. While {\tt see} shows up  with descriptors like \emph{blue, shadow, bright}, and \emph{dark}, the color space of {\tt to look} is more diverse including words like \emph{pale, white, gray, brown, green, blue, yellow}. 

One clear characteristic of the descriptor space of the verb {\tt to look} that {\tt see} does not seem to have is captured by many words of physical and emotional appearance (i.e., \emph{attire, appearance, personality}), as well as a variety of evaluative adjectives such as \emph{good, bad, neat, nice, cool,
foolish, silly, stupid, odd, 
funny, tired, sad}. On the other hand, {\tt to see} tends to occur more with emotion words like \emph{happy, glad, delighted, joy, pleasure, love, sad, hate}.

\subsection{The Auditory Modality: \emph{to Listen} vs. \emph{to Hear}}

Given that only 18\% of the descriptors were shared by the two auditory verbs (i.e., \emph{music, song, tune, radio, heartbeat}), it is the difference between the two spaces that is more interesting. A first clear distinction is shown by the adverbial descriptors. The two verbs differ mostly across their choice of adjectives of intensity and adverbs of manner: {\tt to hear}: (e.g., \emph{loud, distant, muffle, faint}), {\tt to listen}: (e.g., \emph{actively, silently, respectfully, eagerly, patiently, carefully, closely, attentively, intently}). This shows a clear agentive tendency of the verb {\tt to listen}. As with the verbs of vision, the auditory verbs also differ along the static/dynamic aspect. {\tt To listen} occurs more in the context of presentations, lectures, discourses, prayers, sermons, CRMs, while {\tt to hear} is associated more with contexts of courtrooms and judiciary committees. The movement-rich space of {\tt to hear} is also indicated by descriptors like \emph{knock, footsteps, heartbeat, sigh, laughter, scream, shout, cry, explosion, gunfire, shot}.

Unlike {\tt hear}, {\tt to listen} is rich in descriptors denoting empathy, caring, compassion, openness and calmness, with an inclination toward concepts of intuition and instinct. The verb {\tt to hear} is seen more with words of emotion (i.e., \emph{sad, love, sorry, glad, anxious, tired)}, and in particular with adverbs of frequency (\emph{shortly, lately, rarely}), intensity (\emph{loud, faint, distant}) and information gathering (i.e, \emph{first-hand, rumor}).

\subsection{The Tactile Modality: \emph{to Touch} vs. \emph{to Feel}}

\begin{figure*}
  \caption{\label{touch-dandogram} The dandogram generated for the descriptors of {\tt to touch} by the clustering algorithm.}
  \centering
    \includegraphics[width=0.95\textwidth]{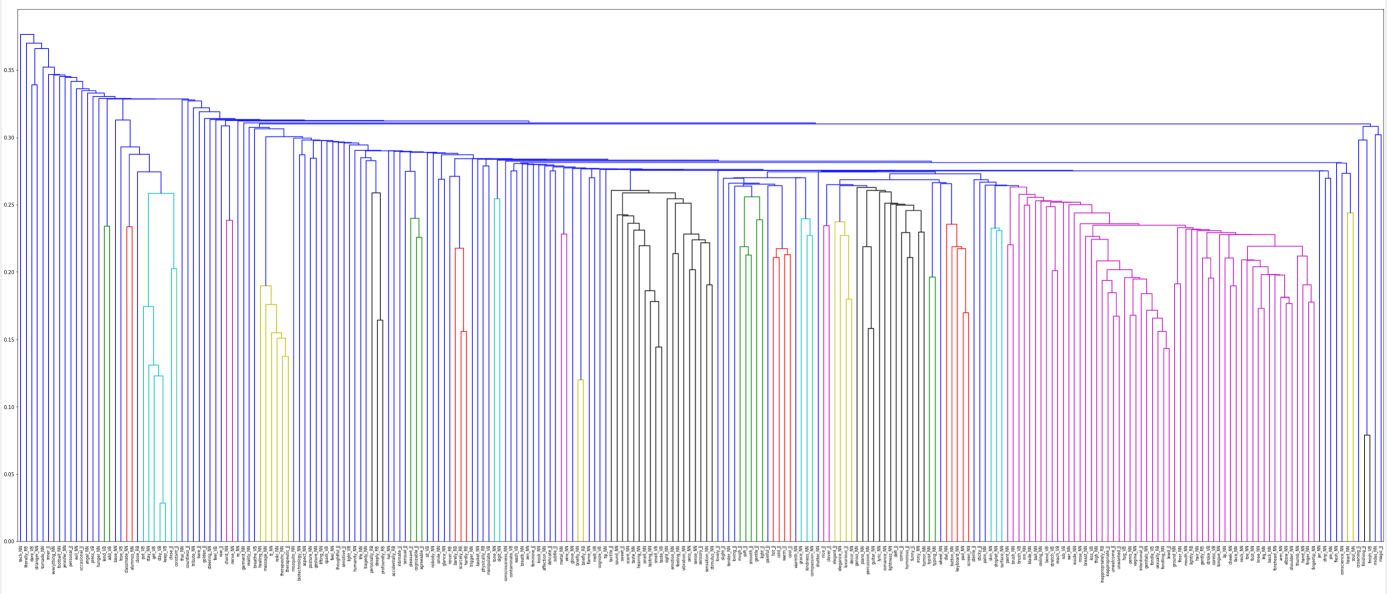}
\end{figure*}

The two tactile verbs shared only 12\% of their descriptors mostly given clusters identifying texture (i.e., \emph{soft, rough, smooth, gentle, light}), temperature (i.e., \emph{hot, warm, cold}), pain (i.e., \emph{painful, sharp, unpleasant}), empathy (\emph{kindness, compassion, sympathy}), as well as body parts (\emph{finger, skin, leg}). 
{\tt Touch} is mostly seen with garments (e.g., \emph{elegant, luxurious, decor}), musical instruments, therapeutic massage, medical tools. However, a clear cluster in each perception verb's space was that of modifiers. While {\tt to touch} occurs with adverbs at the lower end of the scale (i.e., \emph{never, barely, scarcely, hardly}), {\tt to feel} is seen with adverbs and adjectives of high intensity at the other end (i.e., \emph{completely, totally, absolutely}). A special subcluster of {\tt to touch} was captured by sexually-related modifiers like \emph{unwanted, inappropriate}. Due to space constraints we show here only the dandogram generated by the clustering algorithm for the tactile verb {\tt to touch} (Figure \ref{touch-dandogram}).

The largest cluster for the verb {\tt to feel} (55\% of all descriptors) identified mental states with distinct phenomenology related to mind, soul; consciousness, perception; and sensation, thought, emotion.
Given the very large size of this cluster, we ran the clustering algorithm again, thus resulting in more detailed sub-clusters. 
Sub-cluster1 identifies sensations (interoception: \textit{cold, hot, cool, warmth}).
Sub-cluster2 captures mental-state verbs of interpretation (\textit{to believe, think, have an opinion, imagine} - "\textit{He feels he must resign.}"). Sub-cluster3 captures emotional reactions ("\textit{I feel angry.}"; or a wish: "\textit{I feel like taking a walk.}"). 

While our sensory receptors are constantly collecting information (\textit{senses}) from inside and outside of our body, it is ultimately how we \textit{interpret} that information and how we \textit{react} to it that defines our interaction with the world. The emotion reactions sub-cluster maps very well onto well established lists of emotions, like Parrott's emotions \cite{Parrott2001} and Plutchik's Wheel of Emotions \cite{plutchik2001} at various granularity levels:
love/affection/compassion; love/desire, passion; joy/happiness, satisfaction; anger/hatred, frustration, resentment; sadness/sympathy/pity; sadness/shame/guilt/remorse; sadness/depression, despair, unhappiness, grief, sorrow;  sadness/ isolation, loneliness, alienate; sadness/neglect/embarrassment, insecure, disappointment; fear/nervousness; fear/stress, anxiety, isolation, helplessness; happy/respect, value; Angry/distant/numb.
Our clustering algorithm also uncovered novel descriptors like social pressure/commitments, obligations; pressured, bind, obligate, oblige, compel.

\subsection{The Taste and The Smell Modalities}

Previous research has shown that, in identifying odors, we often use either source-based descriptions (e.g., \emph{citrusy}), abstract descriptions (e.g., \emph{musty}),  or  metaphorically  used cross-modal sensory  property (e.g., \emph{light}),  or evaluative/hedonic descriptions (e.g., \emph{pleasant}) 
\cite{CroijmansM15,MajidGLMA18}.

In our experiments, the descriptor spaces generated for the verbs {\tt to taste} and {\tt to smell} shared 28\% of their concepts mostly given clusters identifying many source-based  (i.e., \emph{fishy, smoke, curry, garlic, chlorine, strawberry, chocolate, coffee, spice, cigarette, wine, tea, cheese}), evaluative/hedonic descriptions (i.e., \emph{foul, agreeable, unpleasant, delicious, awful, pleasant, horrible, nasty, wonderful}), as well as generic olfactory terms like \emph{odour, odor, aroma, scent}. There were only a few abstract descriptions (i.e., \emph{earthy, pungent}) and metaphorically  used cross-modal sensory  property (e.g., \emph{slightly, freshly)}, Of all the shared descriptors, 54\% were nouns, 35\% adjectives, 8.33\% verbs, and the rest, adverbs. The verb descriptors in common were \emph{touch, linger, sound, cook, dress, eat}. 

The smell--taste descriptor spaces had unique concepts as well, like very unpleasant smells \emph{putrid, fetid, rotten, mildew} as well as indigestible objects like \emph{perfume, cologne, ammonia} (for {\tt smell}). Unique {\tt taste} descriptors included edible objects like \emph{tarte, appetizer, salsa, broth, honey, mustard} and adjectives \emph{refined, nutritious, crisp, aesthetic}.

\section{Descriptor Spaces across Sensory Modalities}

In this section, we present experimental results of our further investigation of the differences and similarities of our descriptors' spaces along the basic sensory modalities. 
Such distinctions are extremely important when considering interactions between linguistic, conceptual, and perceptual systems. A common assumption of empirical cognitive studies of perceptual words has been that a concept can be experienced through just one perceptual modality. More recently, researchers in cognitive science have empirically derived measures that allow to classify words as unimodal, bimodal, or multimodal, while separately considering the representational strength on each perceptual modality \cite{Lynott2020}.


We thus took advantage of external datasets, like the freely available Lancaster Sensorymotor Norms Dataset of words ranked on different semantic dimensions \cite{Lynott2020}, to date the largest set of semantic norms for English. Specifically, the dataset contains 37,058 single word lemmas (e.g., \emph{bus}) selected from previous similar norm sets, covering the major syntactic categories (e.g., nouns, verbs, prepositions, pronouns, adjectives, adverbs) and a wide range of concepts (e.g., foods, animals, emotions, sports, taboo words, professions, colors). The norms provide a wide lexical coverage with words' lemmas being known by at least 85\% of native English speakers \cite{Lynott2020}. 
Each word is listed with ratings of perceived strength of association with the five basic sensory modalities. 
These ratings were provided by 3,500 English-speaking human subjects via Mechanical Turk on a 1--5 Likert scale capturing the extent to which they 
experience everyday concepts (given by lemma and part of speech) along six sensory modalities -- the Visual, Auditory, Tactile, Interoception, Olfactory, and Gustatory\footnote{'For perceptual strength norming, the subjects were instructed to answer the question: “To what extent do you experience WORD,” where WORD was replaced with each lexical item in the norm dataset in uppercase text. Underneath were six rating scales, one for each of the perceptual modalities considered, labeled “By feeling through touch,” “By hearing,” “By sensations inside your body,” “By smelling,” and “By tasting”; the order of these modalities was randomized for each item list. At the bottom of the screen were a check box labeled “Don’t know the meaning of this word” and a button labeled “Next.” ' (excerpt from Lynott et al, 2020 study \cite{Lynott2020}).}. The rating scales ran from 0 (not experienced at all with that sense modality) to 5 (experienced greatly with that sense modality), or don't-know -- in case the subjects did not know the meaning of the word.

Norm-generating studies involving tens of thousands of words have become increasingly popular across the cognitive sciences particularly due to their ability to provide greater statistical power, reduce experimenter bias in item selection, and increase study reliability \cite{Lynott2020}. 
However, it is also important to mention that perception ratings of such wide coverage lexical lists modeled across various sensory modalities are usually determined in an out-of-context fashion via human subjects questionnaires. Our sensory descriptors, however, were identified in and compared across the textual contexts in which they occurred.

For this component of our study, we were particularly interested in the following perception scores with which the Lancaster project's subjects rated the lexical items:
\begin{enumerate}
    \item {\bf Perceptual Exclusivity} -- a measure of the extent to which a particular concept (i.e., word lemma with corresponding part of speech) is experienced through a single sensory modality. Average exclusivity scores can be expressed as a proportion (as the rating range of lexical items divided by the sum) and extend from 0 (completely multisensorial and experienced equally in all sense modalities) to 1 (completely unisensorial and experienced solely through a single sense modality).
    \item {\bf Average Maximum Strength}. We identified the dominant sense modality of each concept in the Lancaster norms and in our dataset according to which sensory modality has the highest rating (i.e., maximum sensory strength), and labeled the dominant perceptual modality per concept. In case of a tie for the highest rating, we followed Lynott and colleagues' approach and assigned a dominant modality at random from the tied candidates.
\end{enumerate}
For our analysis, we extracted the ratings of those Lancaster norms' concepts that overlapped with the lemmas and parts of speech of our descriptors. The resulting assignment of ratings of descriptors with strengths as dominant modality is shown in Table \ref{sensory-modalities}.
We also report here the exclusivity scores of the eight verbs of perception across the six perceptual modalities. Specifically, Table \ref{perc-exclusivity} shows the average perceptual exclusivity score (and its standard deviation) and the average maximum strength score (with its standard deviation) -- as the rating range of each descriptor divided by the total number of each of the eight sensory spaces (i.e., per perceptual verb, and averaged across modalities). The table also shows the total count for each 300 perception verb lists of descriptors in our perception dataset. 


\begin{table}
\centering
\tiny
\begin{tabular}{lrrrrrrr}
\hline \textbf{Verbs} & \multicolumn{7}{c}{\textbf{Dominant Sensory Modalities}} \\ \cline{2-8} 
\textbf{} & \textbf{A} & \textbf{G} & \textbf{H} & \textbf{I} & \textbf{O} & \textbf{V} & \textbf{NA} \\ \hline
See    & 5.33 & 0.33 & 0.67 & {\bf 14} & 0.67 & {\bf 72} & 7 \\
Look   & 7    & 0.33 & 2.67 & 8.67 & 0.67 & {\bf 77.67} & 3 \\
Hear   & {\bf 39.67} & 1 & 1 & 9.33 & 0.67 & {\bf 42} & 6.33 \\
Listen & {\bf 39.33} & 0 & 0.67 & 12 & 0.33 & {\bf 37.67} & 10 \\
Touch  & 7.67 & 1.67 & {\bf 16.67} & 12 & 1 & {\bf 56.67} & {\bf 43.33} \\
Feel   & {\bf 9.33} & 0.33 & 6 & {\bf 44.33} & 0.33 & {\bf 36.67} & 3 \\
Smell  & 5.33 & {\bf 13.67} & 9.67 & 7 & {\bf 13} & {\bf 48.67} & 2.67 \\
Taste  & 4.33 & {\bf 37} & 3.67 & {\bf 10} & {\bf 5.33} & {\bf 37} & 2.67 \\
\hline
\end{tabular}
\caption{\label{sensory-modalities} The distribution of each 300 verb descriptors with strengths as dominant modality across V (Visual), A (Auditory), H (Haptic or Tactile), I (Interoceptive), O (Olfactory) and G (Gustatory) sensory modalities. NA refers to concepts not found in any of the above categories.}
\end{table}

\begin{table}
\centering
\small
\begin{tabular}{llllll} \hline 
\textbf{Verbs} & \textbf{APE} & \textbf{Std} & \textbf{AMS} & \textbf{Std} & \textbf{Count}  \\ \hline
See   & 0.42 & 0.19 & 3.51 & 0.98 & 279 \\
Look  & 0.41 & 0.18 & 3.46 & 1 & 291  \\
Hear  & 0.46 & 0.14 & 3.37 & 1 & 281 \\
Listen & 0.45 & 0.15 & 3.68 & 0.81 & 270 \\
Touch  & 0.39 & 0.15 & 3.61 & 0.82 & 287 \\
Feel   & 0.35 & 0.13 & 3.23 & 0.88 & 291 \\
Smell  & 0.35 & 0.16 & 3.88 & 0.76 & 292 \\
Taste  & 0.31 & 0.13 & 3.81 & 0.79 & 292 \\
\hline
\end{tabular}
\caption{\label{perc-exclusivity} The average perceptual exclusivity (APE) scores with standard deviation, average maximum strength (AMS) with standard deviation, and the total count for each of the 300 descriptors.}
\end{table}


As the numbers indicate, overall, our sensory descriptors were highly multidimensional (i.e., multisensorial), with average exclusivity scores in the range 0.31 -- 0.46. 
The most multidimensional words per modality in our dataset are (with sensorimotor exclusivity scores): Visual (\emph{everything}: 0.03), Auditory (\emph{anything}: 0.07), Interoceptive (\emph{sense}: 0.55), Haptic (\emph{sensual}: 0.15), Olfactory (\emph{freshly}: 0.18), Gustatory (\emph{nutritious}: 0.13). 
The most unidimensional words per modality are: Visual (\emph{look}: 0.98; \emph{white}: 93), Auditory (\emph{thunder}: 0.8), Interoceptive (\emph{soul}: 0.74), Haptic (\emph{sticky}: 0.57), Olfactory (\emph{stench}: 0.85), Gustatory (\emph{flavor}: 0.57)  given the concept scored strongly on given modality strength but weakly on everything else. It is important to note here that high exclusivity does not necessarily mean high strength. Considering modality exclusivity alone, \emph{rainbow} is strongly visual, while \emph{unbudgeted} is weakly visual, yet both score 100\% on modality exclusivity (meaning, they are unimodally visual concepts), while all other modalities have perceptual strength of zero \cite{Lynott2020}.

Overall, a large percentage of our descriptors across all eight perception verbs considered here were visually dominant (Table~\ref{sensory-modalities}). 
Previous research has shown that Gustatory and Olfactory spaces are highly correlated, as are Visual and Tactile, while Auditory and all other modalities (minus Interoception) were negatively associated (see \cite{Lynott2020}).
In this paper, we take a step further and look at these associations along the Agentive/Experiential dimension across all modalities, as indicated by our set of perceptual verbs. The following subsections provide more details about their semantic spaces.

\subsection{Descriptors of Vision across Sensory Modalities}

Over two thirds of our descriptors of sight were visually dominant - which matches our heatmap (Table \ref{heatmap}), followed by concepts with high interoceptive experience (9-14\%) corresponding to fairly high association with words of feeling, in our datasets, as well. What draws particular attention here is the Vision - Auditory  correlation. While the overall numbers match our observations, our heatmap shows a particular increased overlap in descriptors of {\tt to see} and {\tt to hear}, and less between {\tt to look} and {\tt to hear}. The intersection with the Lancaster dataset descriptors shows insignificant overlap of Vision with Gustatory and Olfactory modalities, yet our dataset indicates a higher tendency of association between {\tt to look} and {\tt to smell}. When it comes to the Tactile sensory space, the Lancaster dataset brings forward a slight increase of {\tt to look} words which is confirmed by our lists of descriptors for Vision (both {\tt to see} and {\tt to look}) and Tactile senses (only {\tt to feel}). This increase also explains the higher numbers of Vision - Interoception words. However, our heatmap shows a higher overlap of {\tt to look} with {\tt to touch, feel, smell}, and moderately with {\tt to listen}. This shows that when we look closely, with intent (i.e., with high agency), we pay more attention to the texture and shapes of objects and the emotions they evoke. The medium-high correlation between {\tt to look} and {\tt to smell} is indicated by words like \emph{flower, mud}.

All these observations indicate that the characteristics of proximity and intent of the descriptor spaces of the verb {\tt to look}, generate experiences that cross sensory modalities, experiences that are more intuitive and personal allowing for a more intuitive, intimate understanding of the world around us, than the one offered by the verb {\tt to see}.

\subsection{Auditory Descriptors across Sensory Modalities}

About 36-42\% of the descriptors of {\tt to hear} and {\tt to listen} had both Auditory and Visual modality interpretation, according to the intersection with the Lancaster norms. This observation is corroborated by our heatmap and the clusters analyzed per perception verb. The Audio - Interoception overlap (9-12\%) is also in line with our observations of mentions of body parts, feeling/emotion words and evaluative modifiers of hearing. The descriptors of {\tt to listen} are more associated with those of Vision ({\tt to look} and {\tt to see}), Tactile ({\tt to feel} and {\tt to touch}), while {\tt to hear} is correlated more with {\tt to see} and {\tt to feel} (e.g., due to the empathic/compassionate descriptors in the {\tt to listen} set). This is mainly due to the importance of the proximity and intent dimensions that seem to separate the Agentive/Experiential verb spaces.

\subsection{Tactile Descriptors across Sensory Modalities}

Over 50\% ({\tt to touch}) and 37\% ({\tt to feel}) words had higher strength of experience with the Lancasters' descriptors of Vision. This is supported by our datasets, with the difference that we noticed a much higher association between the spaces of {\tt to feel} and those of Vision, than those of {\tt to touch}. 44\% of the {\tt to feel} concepts (and only 12\% of {\tt to touch}) are associated with Interoception.
Another rather striking difference comes from the intersection of {\tt to feel} and {\tt to touch} with Gustatory and Olfactory modalities. The intersections with our dataset show a rather high (16\%) overlap of {\tt to feel} and {\tt to touch} with Olfactory descriptors, and about 6\% overlap with Gustatory concepts. 

\subsection{Olfactory and Gustatory Descriptors across Sensory Modalities}

About 49\% and 37\% of the Lancasters' words of Vision overlapped with our descriptors of Olfactory and Gustatory modalities, respectively, and ranked rather low on associations with Tactile, Interoception, and Auditory. This is somewhat different from our observations on our own dataset which show that {\tt to smell} is particularly correlated with {\tt to look} (and somewhat with {\tt to see, touch, feel}). A closer look at the descriptors of {\tt to smell} show subclusters like \emph{salt, breeze, sea, ocean} that are spread across a large number of modalities (Visual, Olfactory, Gustatory, Interoception, and Audio). {\tt To smell} is also well correlated with {\tt to touch} and {\tt to feel} as in the subcluster \emph{feel, keen, soft}. 

Our heatmap also shows that {\tt to taste} has an average association with Vision, but more with the Olfactory and Tactile modalities (and an average strength with {\tt to feel}). This is mainly due to the high number of food-associated words that convey information of texture and temperature, thus giving rise to pleasant feelings.

\section{Discussion and Conclusions}
 
In this paper we reported on a data-driven approach to identifying and organizing the descriptor space of the main perception verbs along the Agentive/Experiential dimension and across six modalities: Vision, Auditory, Tactile, Olfactory, and Gustatory (we added Interoception when compared with the Lancaster sensorimotor norms). In the analysis, we identified differences and similarities of the perceptual descriptors based on qualitative and quantitative differences of the perceptual experience they denote. 

The sensory perceptions transduced by our sense organs are different, yet complementary ways of interacting with and experiencing the world around us. They inform us in different ways with different emotional impact. 
While the perceptual spaces of the experiential verbs {\tt to see, hear, touch} show a more detached, logical way of knowing and learning, their agentive counterparts ({\tt to look, listen, feel}) provide a more intentional as well as more intimate and intuitive way of discovering and interacting. One of the many descriptor examples that stood out in our experiments was the subcluster containing words like \emph{first-hand} and \emph{rumor} in the Auditory perception space, as opposed to that of Vision. 
It has been long believed that knowing the world by seeing should be more believable than by following auditory paths. This shows in day-to-day verbal patterns like \emph{"I see"} vs. \emph{"I hear"} -- that of in-sight vs. hearsay - where hearing something comes across as shady, not less than gossip.

Visual descriptions, on the other hand, are rarely just visual. For instance, based on the results obtained on our large English corpus, Vision and Tactile modalities have a special affinity. In sighted people, Vision (the verb {\tt to see}) seems to provide information about the surrounding world on a grand scale, while {\tt to look} is more proximal, and {\tt to touch} and {\tt to feel} inform on a smaller, more intimate scale. {\tt To touch} and {\tt feel} are also connected with hearing. When we listen, the sound touches us.

It is important to mention that, while our approach is flexible in processing large amounts of data, our research is limited in a number of ways. Our data-driven approach to sensory perception is language and genre independent. Yet, its results apply to English verbs as occurring in our large dataset, and do 
not necessarily generalize to other natural languages. Only a multilingual study would be able to test the similarities and differences to other languages and to human perception in general. Even when it comes to research conclusions about English perception verbs, the results may vary with the dataset and the setting of the model, as the parameter values are set manually or determined empirically. 

The perception verbs considered in this research have different categorization frames that occur across the Agentive/Experiential divide (e.g. {\tt look at} vs. {\tt see}). Such syntactic information is important and can better inform the choice of some of these model parameters like the size of the context window. However, the inclusion of syntactic information is very costly to obtain and use for large data driven approaches like ours. One might think that the analysis for the verb {\tt to look}, for instance, would suffer from having a smaller set of content words being considered (given a small context window). While no large data model identifying the sensory spaces of perception verbs can guarantee the extraction of all the descriptors, we believe that this is not really a problem for us since the verbs of perception we consider do not have long dependencies, and thus, many of their descriptors can be captured in rather small context windows.

This research has a wide number of applications. One that is particularly important to us is how can we best capture the cooperation of human sensory modalities to help people with visual impairment. Sensory compensation refers to the lack or alteration of one sensory modality that could change the distribution of input from the other sensory modalities. Cognitive linguists Landau \& Gleitman \cite{Landau-Gleitman1985} believe that the linguistic contexts in which a verb is used may be a crucial contributing factor for blind children to understand and use verbs such as {\tt to look} and {\tt to see}. A well understood mapping of the linguistic sensorium would be of tremendous benefit in providing experiential analogies with related sensory experiences for the blind. Of particular importance are experiences in the Auditory ({\tt to listen} vs. {\tt to hear}) and Tactile modalities ({\tt to touch} vs. {\tt to feel}) with clear analogies in the Vision space ({\tt to look} vs. {\tt to see}). This study is part of a large-scale project focusing on sensory spaces across the traditional sensory inputs whose purpose is to do just that.

\bibliographystyle{elsarticle-num-names} 
\bibliography{cas-refs}





\end{document}